\documentclass{article}

\usepackage{microtype}
\usepackage{graphicx}
\usepackage{subcaption}
\usepackage{booktabs} 

\usepackage{hyperref}

\usepackage[accepted]{icml2026}

\usepackage{amsmath}
\usepackage{amssymb}
\usepackage{mathtools}
\usepackage{amsthm}

\usepackage[capitalize,noabbrev]{cleveref}

\usepackage{multirow}
\usepackage{makecell}

\theoremstyle{plain}

\theoremstyle{definition}

\theoremstyle{remark}

\usepackage[textsize=tiny]{todonotes}

\icmltitlerunning{How do Humans Process AI-generated Hallucination Contents: a Neuroimaging Study}

\begin{document}

\twocolumn[
  \icmltitle{How do Humans Process AI-generated Hallucination Contents:\\ a Neuroimaging Study}



  \icmlsetsymbol{equal}{*}

  \begin{icmlauthorlist}
    \icmlauthor{Shuqi Zhu}{thu}
    \icmlauthor{Yi Zhong}{thu}
    \icmlauthor{Ziyi Ye}{fdu}
    \icmlauthor{Bangde Du}{thu}
    \icmlauthor{Yujia Zhou}{thu}
    \icmlauthor{Qingyao Ai}{thu}
    \icmlauthor{Yiqun Liu}{thu}
  \end{icmlauthorlist}

  \icmlaffiliation{thu}{Department of Computer Science and Technology, Tsinghua University, Beijing, China}
  \icmlaffiliation{fdu}{Institute of Trustworthy Embodied AI, Fudan University, Shanghai, China}

  \icmlcorrespondingauthor{Ziyi Ye}{zyye@fudan.edu.cn}
  \icmlcorrespondingauthor{Qingyao Ai}{aiqy@tsinghua.edu.cn}

  \icmlkeywords{hallucinations,brain signals,neuroscience,multimodal large language model}

  \vskip 0.3in
]



\printAffiliationsAndNotice{}  

\begin{abstract}

While AI-generated hallucinations pose considerable risks, the underlying cognitive mechanisms by which humans can successfully recognize or be misled by these hallucinations remain unclear.
To address this problem, this paper explores humans' neural dynamics to characterize how the brain processes hallucinated content.
We record EEG signals from 27 participants while they are performing a verification task to judge the correctness of image descriptions generated by a multi-modal large language model~(MLLM).
Based on an averaged event-related potential~(ERP) study, we reveal that multiple cognitive processes, e.g., semantic integration, inferential processing, memory retrieval, and cognitive load, exhibit distinct patterns when humans process hallucinated versus non-hallucinated content.
Notably, neural responses to hallucinations that were misjudged versus correctly judged by human participants showed significant differences. 
This indicates that misjudged AI-generated hallucinations failed to trigger the standard neurocognitive fact verification pathway.
The detailed code can be accessed openly through the url \url{https://github.com/Promise-Z5Q2SQ/EEG-Hallucination}.
\end{abstract}

\section{Introduction}

Over the past few years, AI models have made impressive progress, scaling in size, architecture sophistication, and capability~\cite{minaee2024large,zhao2023survey,wu2023multimodal}.
These advances have enabled them to perform a wide range of tasks, from image captioning and generation~\cite{borji2022generated} to open-ended conversation~\cite{touvron2023llama,xu2026ael} and multi-modal understanding~\cite{qwen2.5-VL,yang2023dawn,guo2026memeyevisualcentricevaluationframework}. 
However, one of their key drawbacks, hallucinations, i.e., the tendency to generate plausible yet factually incorrect content, has become a growing concern. 
A series of studies has shown that such hallucinations can easily mislead humans to make erroneous decisions~\cite{sun2024ai} in critical domains such as healthcare, law, and finance.

To address hallucinations in AI models, researchers have attempted to understand the origins of AI hallucinations, especially in multi-modal large language models (MLLMs) and large language models LLMs~\cite{ji2023survey,huang2025survey}.
Much of existing research has investigated hallucinations from the perspective of the model, i.e., exploring how aspects such as training data, prompt design, decoding or sampling strategies, and internal uncertainty or confidence measures contribute to the occurrence of hallucinations~\cite{maynez2020faithfulness}. 
Based on the investigations, researchers further work on detecting hallucinated content automatically and further deriving mechanisms to generate factually consistent content.
For example, \cite{lewis2020retrieval} introduced retrieval-augmented generation to augment the LLMs with factual knowledge.
\citet{farquhar2024detecting} proposes using several verification steps during generation to check the consistency of the generated content. 

On the other hand, the effects of AI hallucinations on humans have been extensively studied
~\cite{zhai2024effects,kim2025fostering}. 
Most of these studies are based on users' explicit, post-hoc judgments and behaviors during their interactions with AI hallucinations~\cite{barros2025think}. 
For example, \citet{klingbeil2024trust} conducted a user study examining AI hallucinations with varying levels of fluency and presentation tone.
It reveals that humans are prone to being misled by content characterized by high fluency and professionalism.
This indicates that AI-generated hallucinations vary significantly in their deceptiveness, resulting in different levels of risk to human cognition and decision-making.


However, few studies have examined, from a neuroscience perspective, how human brain activity patterns differ when viewing hallucinated versus non-hallucinated content generated by AI models.
Existing theory has segmented human perception of information content into several stages based on humans' EEG signals, progressing from early sensory encoding and attentional allocation to higher-order semantic integration and memory retrieval~\citep{luck2014introduction}.
Building on this theoretical framework, a critical question arises: at which processing stage does the detection of AI hallucinations succeed or fail? 
Investigating these temporal dynamics is essential to understanding why humans are susceptible to ``plausible but incorrect'' AI hallucinations.

To fill this gap, we investigate the neurological mechanisms by which humans recognize hallucinations and gain insight into the development of MLLMs. 
Specifically, we have the following research questions.

\begin{itemize}
\item \textbf{RQ1.} Do neural signals exhibit significant differences across distinct temporal stages when participants view hallucinated versus non-hallucinated contents?

\item \textbf{RQ2.} If yes, is this difference modulated by whether participants correctly recognize the hallucination content?

\item \textbf{RQ3.} Can we predict whether AI-generated content contains hallucinations based on human neural signals?
\end{itemize}

To address these research questions, we collected EEG data from 27 participants. 
Each participant viewed textual stimuli generated by an MLLM that included both hallucinated and non-hallucinated content. 
Participants were asked to judge whether the textual stimuli matched the image content (i.e., whether they recognized any hallucinations). 
On the basis of this paradigm, we conducted averaged event-related potential~(ERP) analyses~(detailed in Section~\ref{sec: ERP Analysis}).
Based on the EEG/ERP methods, we obtain a series of findings about the underlying mechanisms when humans are perceiving AI-generated hallucinations.
The ERP analysis reveals a significant difference in brain signals when participants are processing hallucination words and non-hallucination words. 
And it suggests that multiple cognitive processes, such as semantic-thematic integration, inferential processing, memory retrieval, and cognitive loading, are engaged in hallucination recognition. 


However, we also noticed a significant difference between the brain signals when the hallucination words are misjudged and correctly judged by participants.
In-depth analysis across different temporal stages suggests that misjudged AI hallucinations exhibit different neurocognitive patterns, specifically affecting humans' allocation of attention and inferential reasoning.
Inspired by those analyses, we further conduct a prediction experiment and show that EEG signals can be used to predict, at both the word-level and sentence-level, whether content contains hallucinations. 
However, this prediction is not as effective for instances where human subjects fail to correctly identify the hallucinations.
This indicates that deceptive AI hallucinations may deceive humans at both neural and behavioral levels. 

\section{Related Work}

\subsection{Hallucination in LLMs}
Hallucination in LLMs denotes fluent but factually incorrect outputs~\citep{ji2023survey,pang2026steering}. 
Prior work separates intrinsic drivers from extrinsic causes and advances two main strands: detection and mitigation. 
For detection, post-hoc verification with retrieval/KBs checks factuality in knowledge-intensive tasks~\citep{lewis2020retrieval}, while black-box consistency methods flag unstable generations without instrumenting the model~\citep{manakul2023selfcheckgpt}. 
AMBER offers an LLM-free, type-controlled benchmark, and object-level studies document captioning-specific hallucinations~\citep{wang2023amber,rohrbach2018object}.
Complementary strategies include self-critique and verifier pipelines to reject dubious claims and improved cross-modal alignment in MLLMs to curb object and attribute hallucinations~\citep{manakul2023selfcheckgpt,rohrbach2018object}. 
Despite progress, most evaluations remain outcome-based, leaving open when and how humans neurally register hallucinations. 

\subsection{Neuroscience \& AI}

Event-related potentials (ERPs) provide time-resolved markers of cognitive processing~\citep{luck2014introduction}.
N100 and P200 index perceptual and attentional allocation; P300 relates to task-relevant salience; N400 tracks semantic integration and expectancy violations; and P600 reflects late reanalysis and monitoring, including “semantic P600” effects~\citep{bornkessel2008alternative}. 
Computational–neuroscience links show partial convergence between language-model representations and brain activity, and surprisal robustly correlates with N400 amplitude in sentence comprehension~\citep{schrimpf2021neural,frank2013word,ye2025generative}. 
Beyond language, images are widely regarded as strong cognitive stimuli, and previous EEG studies have demonstrated promising performance in image-related tasks such as visual stimulus classification and image reconstruction from neural signals~\citep{zhu2024crosspt,zhu2025brain}. 
Furthermore, EEG studies of short-video polarization show that cognitive impact may remain invisible in surface behaviors while still being measurable in neural activity, and that EEG features can predict exposure to polarized content~\citep{du2025eeg,du2025understanding}.
Building on these insights, we compare ERPs elicited by hallucinated versus non-hallucinated words and condition effects on recognition.
\section{Data Collection}

\begin{figure*}[t]
  \centering
  \includegraphics[width=.95\linewidth]{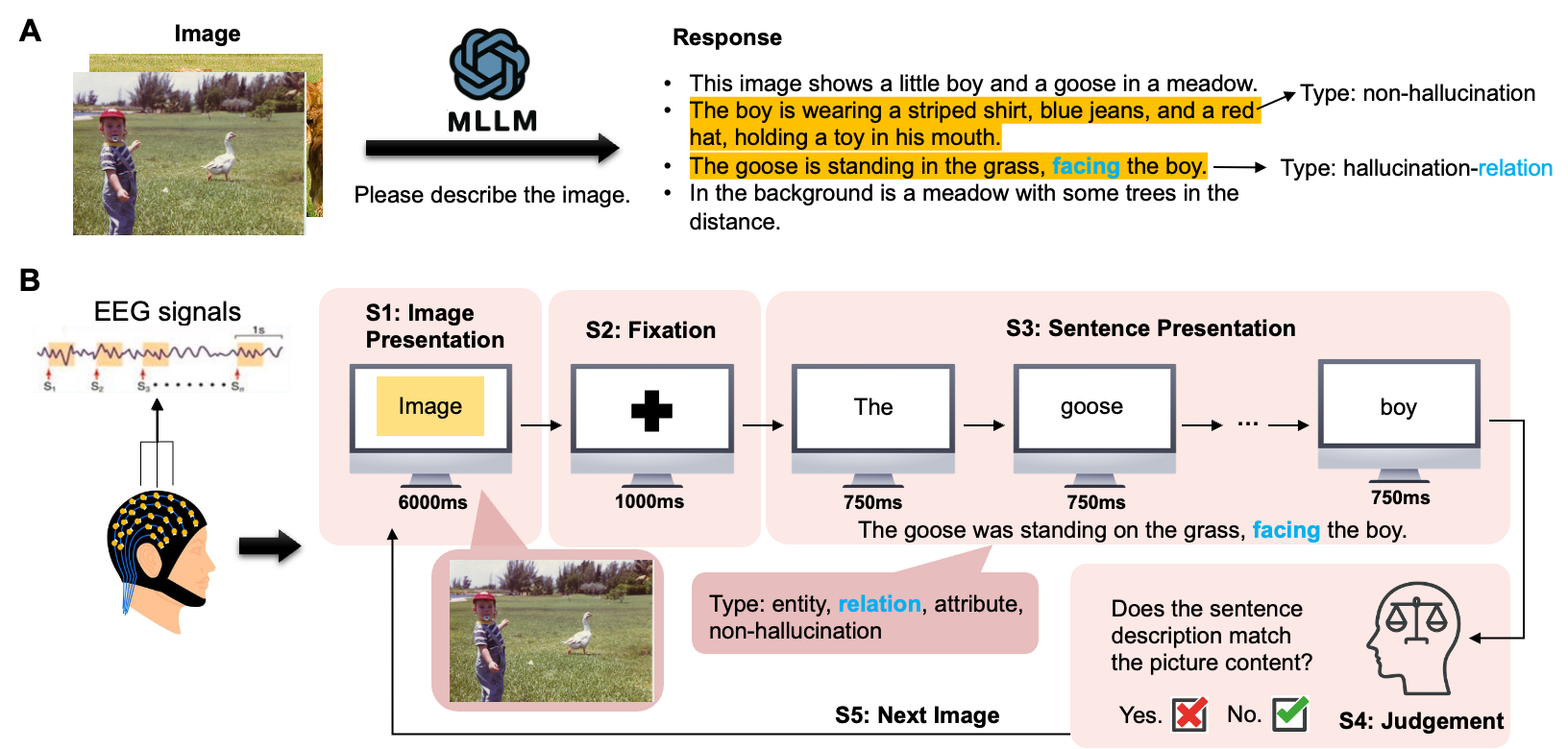}
  \caption{The overall procedure of our data collection. A) The procedure of stimulus selection. B) The experimental trial flow consists of five stages: presenting an image (S1), showing a fixation cross (S2), displaying a sentence word-by-word (S3), the participant making a judgment about the sentence’s match to the image (S4), and finally proceeding to the next image (S5).
}
  \label{fig_procedure}
\end{figure*}

In this section, we describe the collection of EEG and behavioral data from 27 participants while they completed the multi-modal QA task designed for hallucination recognition. 

\subsection{Participants}

A total of 27 volunteers were recruited for this study, comprising 11 males and 16 females, aged between 19 and 30 years (with an average of 24). 
The sample comprised mostly college students, but also included several members of the general public, ensuring some diversity beyond the academic population. 
The participants represented a range of disciplines (e.g., computer science, mechanical engineering, chemistry, and environmental engineering), spanning undergraduate to postgraduate levels.
Each individual completed the full experiment in approximately 1.5 hours, including 30 minutes for equipment setup and task instructions. 
Prior to participation, all individuals were informed that their time would be compensated at a rate equivalent to US\$11.8 per hour, contingent upon their completion of the study, to ensure the quality of the data collected for the study.

\subsection{Task preparation}

To minimize bias arising from participants’ varying disciplinary backgrounds, we deliberately adopted a multimodal QA task that demands minimal prerequisite knowledge. 
Our approach draws on the AMBER benchmark~\cite{wang2023amber}—a multi‑dimensional, LLM‑free evaluation dataset for hallucination in MLLMs—which comprises 1,004 images derived from the MSCOCO~\cite{lin2014microsoft} and includes detailed annotations on three types of hallucinations: entity, attribute, and relation.
Based on this benchmark, we generated responses to generative-style prompts for each image via an MLLM. 
The MLLM we chose in our study is Qwen2.5-VL-3B-Instruct~\cite{qwen2.5-VL, Qwen2VL}.
Leveraging both the original hallucination annotations provided by AMBER and our own manual verification, we manually selected 60 image–response pairs where the Qwen2.5-VL-3B-Instruct generated hallucinatory content. 
For each response, we carefully and manually screened the generated content, selecting one sentence that clearly contained hallucination and one sentence verified to be hallucination-free, thereby constructing a set of strictly controlled and balanced stimuli for EEG testing.
To ensure that each sentence we selected is not an illusion in itself, we use GPT4-HDM method~\cite{su2024unsupervised} and input each sentence into GPT4 to let it judge whether it violates the common sense of the real world.
See Appendix~\ref{sec_verification} for more details.
Following AMBER’s taxonomy of hallucination types, we categorized the stimuli as follows: 27 entity-related, 13 relation-related, and 26 attribute-related (attribute category further subdivided into action (5), count (11), and state (10)). 
Representative cases and the full selection criteria are documented in our publicly accessible code repository.
The procedure of stimuli selection is shown in Fig~\ref{fig_procedure} A.

\subsection{Procedure}

Before the main trials, participants first completed an entry questionnaire and signed informed consent regarding privacy and data security. 
They then received detailed instructions explaining the primary tasks and the operational procedures, and were explicitly informed that they retained the right to withdraw from the study at any time without consequence. 
Following orientation, participants carried out a series of training trials intended to help them become familiar with the formal experiment’s flow. 
Each participant was also asked to select a random seed before the experiment began, which was used to randomize the order of stimulus categories in order to ensure that across participants, each hallucination type and image condition would be fairly and evenly presented.

Once these preparatory steps were finished, each trial proceeded through stages S1 through S5 in sequence, as shown in Fig~\ref{fig_procedure} B. 
In S1 (Image Presentation), an image is shown for 6000 milliseconds while participants are told to view it attentively, knowing there will be a later match judgment. 
In S2 (Fixation), a central fixation cross is displayed for 1000 milliseconds to orient and stabilize visual attention. 
In S3 (Sentence Presentation), a sentence description appears word by word: the first word (e.g., “The”) is shown for 750 milliseconds~(\cite{ye2022towards}), followed by each subsequent word for the same duration; the sentence may either contain a hallucination of a different type or be non-hallucinated. 
After the full description, in S4 (Judgement) participants are asked whether the sentence matches the image content via a binary choice (Yes or No), responding using key presses. 
Finally, S5 introduces the next trial and cycles back to S1 when participants press the space key.
During the entire experiment, we continuously recorded EEG signals from each participant. 
Using event triggers, we logged the onset times of all key stimulus events, so each segment of EEG could be aligned precisely to the relevant experimental stage. 
In addition, for every single sentence shown, we recorded the participant’s judgment (Yes/No) about whether the description matched the image.

\subsection{Apparatus}

The stimuli are presented on a desktop computer that has a 27-inch monitor with a resolution of $2560 \times 1440$ pixels and a refresh rate of 60 Hz. 
Participants are required to use the keyboard to interact with the platform. EEG signals are captured and amplified using a Scan NuAmps Express system (Compumedics Ltd., VIC, Australia) and a 64-channel Quik-Cap (Compumedical NeuroScan). 
A laptop computer functions as a server to record EEG signals and triggers using Curry8 software. 
Throughout the experiment, electrode-scalp impedance is maintained under $50 k\Omega$, and the sampling rate is set at 1000 Hz.

\subsection{Pilot Study}

A pilot study can ensure the correctness of the overall experimental process and the reliability of the acquisition equipment. 
We conduct a pilot study on three partici-pants whose data are not included in the final analysis to determine hyperparameters of the experimental presenta-tion and design.
We also conducted a power analysis based on pilot data, which indicated that 19 participants are sufficient to achieve 80\% statistical power at $\alpha = 0.05$. 
To ensure robustness, we recruited 27 participants, thereby providing sufficient statistical power to detect the reported effects.

\section{Result Analysis}

In this section, we employ ERP analysis techniques to investigate how brain signal patterns differ when participants view hallucination versus non-hallucination words, and synthesize these findings to outline the neural mechanisms by which humans recognize hallucinations.

\subsection{Statistic Analysis}

Across the full set of 120 trials per participant, on average, participants answered 101.14±6.53 items correctly, yielding a mean overall accuracy of 84.29\%. 
Considering hallucination categories, the mean recognition accuracy by type was relation: 90.88\%, entity: 89.30\%, and attribute: 86.18\%. 
Statistical tests reveal that the accuracy across all hallucination categories does not differ significantly.
While the accuracies across these categories are fairly similar, the relation type had the highest performance and the attribute type the lowest, suggesting that relation-based hallucinations may be easier for participants to detect, whereas attribute-level hallucinations pose greater detection difficulty.

\subsection{Preprocess}
\label{sec_prepocess}

We preprocess the EEG data using several steps: first, we re-reference all recorded signals offline using the linked-mastoids method to reduce reference bias~(\cite{yao2019reference}); second, we apply notch, high-pass, and low-pass filters to eliminate environmental interference, slow voltage drift, and high-frequency noise respectively; third, we extract epochs of interest and compute their averages to obtain ERP waveforms. 
The epochs are defined from 200 ms before the presentation of each stimulus word to 800 ms after, covering the expected time window for relevant neural responses.

\subsection{ERP Analysis}
\label{sec: ERP Analysis}

ERP refers to brain voltages that are time-locked to specific events and reflect neural responses elicited by those events~\cite{blackwood1990cognitive}. 
One of its key advantages is the high temporal resolution it offers, and the sequence of ERP peaks provides precise insight into rapid neural processing stages~(\cite{luck2000event}). 
ERP components are evoked amplitude in different post-stimulus time windows, e.g., N100, N400 (negative waves within 100ms, 400 ms), and P200, and P600 (positive waves within 200 ms, 600 ms). 
These standard ERP markers index different cognitive operations~\cite{luck2000event}. 
In our analysis, we employ conventional ERP-processing procedures including signal preprocessing, defining time windows of interest, and specifying regions of interest (ROIs) for comparing conditions~\cite{ye2022towards, zhu2024comparing, ye2024relevance}.
To disentangle different ERP components, we partitioned the extracted time interval into several distinct time windows based on the approach introduced by \citet{lehmann1980reference}. 
Their method identifies evoked scalp potential components by examining both their latency and their topographic pattern. 
We computed the Global Field Power over the 50-750 ms post-stimulus interval, and delineated time windows around the peaks, as shown in Table~\ref{tab_erp_anova}.

\begin{table*}[t]
\caption{The statistical significance test results for different ERP components across brain regions. We use the repeated measures ANOVA test and adopt post-hoc pair-wise comparisons with FDR correction. *  and ** indicate statistical significance at a level of p\textless0.05, p\textless0.001, respectively.} 
\label{tab_erp_anova}
\begin{center}
\begin{tabular}{cccc}
\toprule
Time window                  & ROI                                                                                                 &RM-ANOVA test&Post-hoc test\\
\midrule
\multirow{2}{*}{50--120 ms}& r-temporal, parietal                                                                               &* &HalluCorrect \textgreater NoHallu *\\
 & occipital& *&HalluCorrect \textgreater HalluWrong *\\
\midrule
\multirow{3}{*}{120--280 ms}& pre-frontal, occipital&* &HalluCorrect \textgreater NoHallu *   \\
 & frontal, central, l-temporal&* &HalluCorrect \textgreater NoHallu **\\
 & central, parietal& *&HalluCorrect \textgreater HalluWrong *\\
\midrule                            
\multirow{2}{*}{280--550 ms}& r-temporal&* &HalluCorrect \textgreater NoHallu *\\
 & central&* &HalluCorrect \textgreater NoHallu **\\
\midrule                             
\multirow{3}{*}{550--750 ms}& pre-frontal, frontal, l-temporal,r-temporal,  occipital&* &HalluCorrect \textgreater NoHallu *   \\
 & parietal& *&HalluCorrect \textgreater HalluWrong *\\
                             & central                                                                                             &** &HalluCorrect \textgreater NoHallu **  \\
\bottomrule                             
\end{tabular}   
\end{center}
\end{table*}

\begin{figure*}[t]
  \centering
  \includegraphics[width=\linewidth]{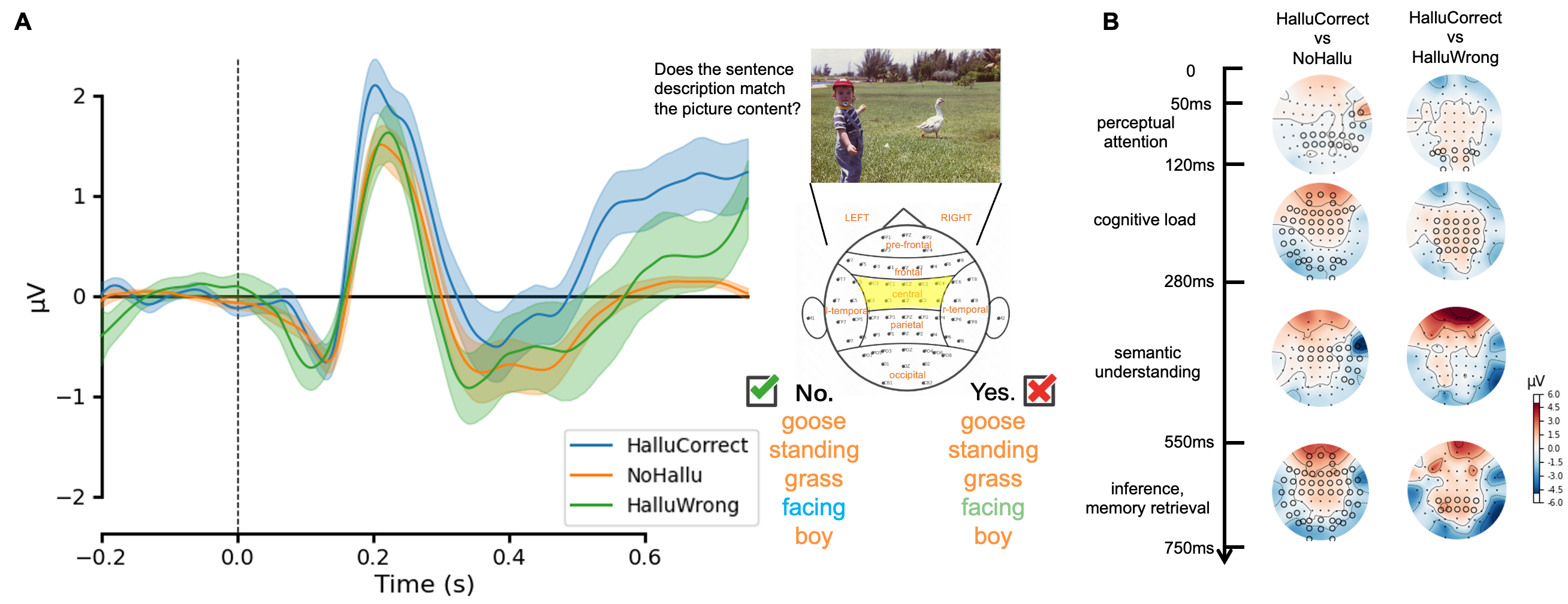}
  \caption{A) Comparison of ERP waveforms elicited by different stimulus word types in the central brain region, with shaded areas indicating the 95\% confidence intervals. B) Time-resolved topographic difference maps comparing HalluCorrect with NoHallu and HalluWrong words, respectively; highlighted electrodes denote brain regions showing significant effects in the post-hoc analysis.}
  \label{fig_erp}
\end{figure*}

To facilitate subsequent analyses, we partitioned the EEG data according to both the stimulus word type and participants’ recognition performance. 
We defined three categories: \textbf{HalluCorrect} for hallucination words that the participant correctly recognized as hallucinated; \textbf{NoHallu} for non-hallucination words correctly identified as non-hallucinated; and \textbf{HalluWrong} for hallucination words which participants failed to recognize (i.e., words that were in fact hallucinations but were judged as non-hallucinations).
We plot the grand-average ERP waveforms for different stimulus word types in central brain region in Figure~\ref{fig_erp}. 
We segment the electrodes into seven brain regions according to their placement on the brain topography shown in Figure~\ref{fig_erp}. 
We applied a repeated-measures ANOVA in a fixed time window for each brain region, followed by post-hoc pair-wise comparisons with FDR correction.
The statistical findings for the various time windows and regions of interest are presented in Table~\ref{tab_erp_anova}. 
Below, we discuss the characteristic features of each component and its potential functional roles.

N100 is an early component in the time window around 100 ms (50–120 ms). 
We employ the repeated measures ANOVA method and discover significant differences between the grand-averaged N100 component in r-temporal (F[1,26]=4.615, p\textless0.05, $\eta_p^2$=0.151), parietal (F[1,26]=4.872, p\textless0.05, $\eta_p^2$=0.158), and occipital (F[1,26]=8.271, p\textless0.05, $\eta_p^2$=0.233).
The N100 component is typically interpreted as reflecting very early visual perceptual processing, especially for low-level visual features~\cite{yang2022event}. 
It often shows maximal expression in occipito-parietal regions. 
Recent work also indicates that the amplitude of N100 is closely linked with attentional allocation.
Larger N100 amplitudes have been observed when stimuli draw more attention, or when perceptual systems are required to allocate greater resources to processing salient or unexpected input~\cite{thornton2007selective, rutman2010early}.
The results of post-hoc test indicate that during the recognition of HalluCorrect words, participants show enhanced N100 responses compared to NoHallu and HalluWrong words. 
This suggests that the process of correctly identifying hallucination words recruits attention very early and imposes a higher cognitive load on the perceptual system, even before later semantic processing stages.

P200 is the dominant component in the time window around 200 ms (120–280 ms).
RM-ANOVA reveals the significant differences between grand-averaged P300 component in pre-frontal (F[1,26]=8.575, p\textless0.05, $\eta_p^2$=0.248), occipital (F[1,26]=11.246, p\textless0.05, $\eta_p^2$=0.302), and frontal (F[1,26]=18.226, p\textless0.001, $\eta_p^2$=0.412), central (F[1,26]=15.311, p\textless0.001, $\eta_p^2$=0.371), l-temporal (F[1,26]=14.037, p\textless0.001, $\eta_p^2$=0.351), and central (F[1,26]=11.285, p\textless0.05, $\eta_p^2$=0.324), parietal (F[1,26]=12.039, p\textless0.05, $\eta_p^2$=0.336). 
The P200 component is generally understood to reflect early attentional engagement and decision-related processing. 
It has been associated with novelty detection, stimulus complexity, and perceptual salience, such that more complex or unexpected stimuli elicit larger P200 amplitudes~\cite{ghani2020erp}. 
Empirical work shows that P200 amplitude tends to increase with attentional load and with stimuli that violate perceptual or contextual expectations~\cite{kemp2009fronto, polich2007updating}.
We observe enhanced P200 responses for HalluCorrect words compared to NoHallu and HalluWrong words. 
This suggests that hallucination words impose greater demands on early stimulus selection.
The processing system flags such words as perceptually or lexically salient, because they diverge from semantic expectation or evoke conflict. 

N400 component is evoked around 400 ms after the stimulus (280-550 ms).
Significant differences are found in central (F[1,26]=16.442, p\textless0.001, $\eta_p^2$=0.387), r-temporal (F[1,26]=7.929, p\textless0.05, $\eta_p^2$=0.234). 
The N400 component is widely understood to index the access and integration of semantic information.
It typically reaches its maximal amplitude at centro-parietal electrode sites, reflecting the brain’s effort to reconcile a word’s meaning with its broader context. 
Empirical findings show that less predictable or semantically incongruent words evoke larger N400 responses, consistent with the idea that the N400 is sensitive to violations of expectation and relates to retrieval from semantic memory~\cite{lau2008cortical,lindborg2023semantic,michaelov2022so}.
HalluCorrect words conflict with the visual context, hence they elicit greater N400 amplitudes than NoHallu words. 
This suggests when the descriptive content generated by the model diverges from what is visually present or semantically expected, the cost of semantic integration increases.
We also note that the post-hoc analysis does not reveal significant differences between HalluCorrect and HalluWrong words. 
This may suggest that even when participants make incorrect judgments, they still subconsciously register semantic conflict. 
However, due to limitations in perceptual processing (e.g., N100, P200) and higher-level inferential or reanalysis processes (e.g., P600), they ultimately arrive at an incorrect decision.

P600 waveform mainly appears in the time window around 600 ms (550-750 ms).
Through ANOVA, we find significant differences between grand-averaged P600 component in pre-frontal (F[1,26]=5.874, p\textless0.05, $\eta_p^2$=0.184), frontal (F[1,26]=7.268, p\textless0.05, $\eta_p^2$=0.218), l-temporal (F[1,26]=7.517, p\textless0.05, $\eta_p^2$=0.224),  r-temporal (F[1,26]=8.093, p\textless0.05, $\eta_p^2$=0.237), occipitial (F[1,26]=6.561, p\textless0.05, $\eta_p^2$=0.202), and central (F[1,26]=31.558, p\textless0.001, $\eta_p^2$=0.548), and parietal (F[1,26]=15.142, p\textless0.001, $\eta_p^2$=0.0.368).
The P600 (or late positivity) component is classically implicated in sentence processing tasks and shows its strongest responses at centro-parietal electrode sites. 
Originally, the P600 was discovered as an index of syntactic reanalysis and repair, reflecting efforts to restructure or repair comprehension~\cite{seyednozadi2021functional}. 
More recently, research has shown that even in sentences that are grammatically correct, semantic conflict or non-typicality can also provoke a P600, which is referred to as the “semantic P600”~\cite{bornkessel2008alternative,brouwer2012getting}.
In the context of hallucination recognition, once a word is identified as semantically hallucinatory, participants engage controlled reanalysis and decision/monitoring processes. 
These processes recruit language-time systems and posterior integration networks, consistent with the late positivity seen in P600. 
Thus, correct detection of hallucination involves not only early sensory/attentional and semantic mismatch stages, but also later re-evaluation and integration when the linguistic input conflicts with perceptual or expectation-based models.



Across all examined components and brain regions in the post-hoc test, none of the comparisons between HalluWrong and NoHallu words reach statistical significance. 
Full test statistics for each time window and ROI are provided in the Appendix~\ref{appendix_erp}.
Importantly, we conducted a post-hoc sensitivity analysis to evaluate the detectability of these effects under our design. 
It shows that our sample size provides approximately 80\% power ($\alpha$=0.05) to detect medium-to-large effects ($d_z\geq$0.462).
This indicates that there exists no medium-to-large ERP effects when comparing HalluWrong versus NoHallu words.
The ERP analysis across different hallucination types did not yield statistically significant results and is therefore not reported in this paper.

\subsection{Discussion}

Overall, our findings advance understanding of the neural mechanisms by which humans recognize hallucinated content generated by MLLMs. 
The ERP results clearly show that various cognitive processes are engaged at extremely fine temporal scales.
Specifically, differences in early perceptual attention and cognitive load (P200/N100), semantic-thematic understanding (N400), inferential processing, and memory retrieval (P600) mechanisms underlie successful hallucination recognition~(addressing \textbf{RQ1}). 
These observations are consistent with prior studies of comprehension mechanisms, which posit that unexpected or incongruent input requires more effortful retrieval and integration of memory and draws upon prediction error and expectancy effects~\citep{zhu2024comparing}.

It is worth noting that our results align with, and extend, findings from previous ERP studies. For example,\cite{ye2022towards} explored the neural mechanisms underlying reading comprehension, and \cite{pinkosova2022revisiting} investigated relevance judgments. 
They both reported that answer words (or words highly relevant to the task) elicit larger ERP amplitudes compared to ordinary or low-relevance words. 
These patterns suggest that when a stimulus is more directly tied to achieving the experimental goal, participants tend to allocate more attentional and cognitive resources to those items.
By analogy, our results suggest that participants in our user study tended to devote more resources to items that were more directly aligned with achieving the experimental task goal, i.e., detection of hallucinated content. 
While the precise mechanisms driving this attentional allocation remain beyond the scope of this paper, they represent an intriguing avenue for future research.

On the other hand, when participants failed to detect the hallucinated content, we did not observe the typical neural activity associated with anomaly detection. 
This may suggest that, due to the high linguistic fluency and contextual coherence, hallucinations produced by advanced models can not successfully trigger cognitive mechanisms related to fact verification.
This may lead to the formation of humans' false beliefs.
In the task we designed, this process primarily operates through attention allocation and inferential processing.
This pattern implies that recognition (or conscious awareness) is a key trigger for abnormal neural responses.
Mere exposure to hallucinated content does not suffice to induce the enhanced ERP effects~(addressing \textbf{RQ2}). 
This aspect distinguishes the present task from many prior word-recognition or anomaly detection tasks.

The findings from our study offer valuable insights:
\textbf{(1)~AI model design:} We reveal that human recognition of hallucinatory information encompasses distinct stages, including attentional engagement, semantic matching, and memory retrieval.
These insights could guide the design of more robust hallucination detection and mitigation systems.
For instance, MLLMs might benefit from a cooperation of modules that do not rely solely on passive prediction but are actively ``aware'' of deviations. 
\textbf{(2)~AI impact on Human:}
Existing research aimed at mitigating AI hallucination often misses important human factors.
From our study, we reveal that the cognitive risk of hallucination content depends heavily on whether humans can successfully recognize it.
Therefore, reducing the cognitive risks associated with such hallucinations constitutes a key research direction.
\textbf{(3)~Human-AI Interaction:} We observe that conscious awareness~(e.g., P200) is essential for triggering anomalous neural responses.
This suggests that user interfaces with active interventions may help reduce the risk of users accepting hallucinated content uncritically. 
Design strategies should emphasize helping users notice and identify when content seems incongruent.

\section{Prediction Experiments}

To explore whether EEG signals can act as signals to predict whether the content generated by an MLLM contains hallucinations, we conducted word-level and sentence-level prediction experiments on the dataset we collected. 
In this section, we detail the procedures and results of experiments.



\begin{table*}[t]
\caption{The classification results of word-level and sentence-level prediction. Best results are in \textbf{Bold}. †/* indicates the result is significantly different with p-value\textless0.05 compared to the best model and random, respectively.}
\label{tab_experiments}
\begin{center}
\begin{tabular}{cccccc}
\toprule
 \multirow{2}{*}{\textbf{Settings}}&\multirow{2}{*}{\textbf{Models}} & \multicolumn{2}{c}{\textbf{Word-level}} & \multicolumn{2}{c}{\textbf{Sentence-level}} \\
                         && $\text{AUC}_\text{within}$    & $\text{AUC}_\text{cross}$    & $\text{AUC}_\text{within}$      & $\text{AUC}_\text{cross}$      \\
\midrule
 \multirow{3}{*}{\makecell[c]{HalluCorrect\\vs\\NoHallu}}&SVM                     &     \textbf{0.9393}*&   \textbf{0.8631}*&       0.9601†*&     0.9494†*\\
 &GBDT                    &     0.9190†*&   0.8362†*&       0.9647*&     0.8531†*\\
 &attention               &     0.9113†*&   0.7955†*&       \textbf{0.9673}*&     \textbf{0.9846}*\\
 \midrule
 \multirow{3}{*}{\makecell[c]{HalluWrong\\vs\\NoHallu}}&SVM                     &     0.5330&   0.5120&       0.4935&     0.5457\\
 &GBDT                    &     0.4935&   0.5217&       0.4870&     0.5392\\
 &attention               &     0.5151&   0.5034&       0.4966&     0.5684\\
 \midrule
 \multirow{3}{*}{\makecell[c]{HalluCorrect\\vs\\HalluWrong}}&SVM                     &     0.5469&   0.4600&       0.4828&     0.4718\\
 &GBDT                    &     0.5316&   0.4748&       0.4897&     0.4728\\
 &attention               &     0.4894&   0.5410&       0.5071&     0.4952\\

\bottomrule
\end{tabular}
\end{center}
\end{table*}

\subsection{Experimental Setup}

\paragraph{Task Definition} We formalize the prediction task as follows. Let a stimulus sentence contain $l$ words, and for each word, we extract EEG features during its presentation. We denote the sequence of word-level EEG features $X = [x_1, x_2, \dots, x_l] \in \mathbb{R}^{l \times d}$ as input, where $d$ is the dimension of the extracted features.
The model produces two outputs: word-level prediction $y_w = [y_{w,1}, y_{w,2}, \dots, y_{w,l}] \in \{0,1\}^l$ and sentence-level prediction $y_s \in \{0,1\}$. For evaluation, we selected AUC as a metric.

\paragraph{Feature Selection} To build input features for our prediction models, we combined Frequency-Band-based Features (FBFs) with Event-Related Potential-based Features (ERPFs). 
FBFs capture global spectral information, while ERPFs focus on specific, behaviorally relevant time windows indicated by our ERP analyses.
We selected four brain regions (central, l-temporal, r-temporal, and occipital) that consistently showed strong effects in our significance tests in the previous section.
We selected a set of time points within those time window and divided each into five equal segments.
For each of those four regions, we computed differential entropy for five standard EEG frequency bands.
Differential entropy is widely used to quantify complexity in EEG signals, and has been shown to be effective for classification tasks such as emotion recognition~\cite{chen2019feature,duan2013differential}.
We concatenated them to create a 760-dimensional input vector.

\paragraph{Data splitting strategies} To examine whether the model's performance is consistent across different participants and whether it generalizes robustly across varying participant data distributions, we employed two data splitting strategies. 
Within-subject paradigm means for each participant individually, we perform ten-fold cross-validation, and then average performance across folds. 
Across-subject paradigm means that we hold out one participant’s data as the test set and train the model on the other 26 participants’ data. 

\paragraph{Model selection} We selected support vector machine (SVM), gradient boosting decision tree (GBDT), and an attention-based model. 
Our rationale for not selecting more complex or highly specialized neural architectures is twofold: 1) this task is novel, and to our knowledge, no dedicated model has previously been designed specifically; 
2) our goal in this work is to demonstrate the effectiveness of EEG as an implicit feedback signal for predicting hallucinated content.
More sophisticated architectures to push maximal performance remain a promising direction for future work.
For more model and training details, please refer to the code and Appendix~\ref{sec_model}.

\subsection{Results}

Table~\ref{tab_experiments} presents the results of the word-level and sentence-level prediction classification.
When comparing HalluCorrect and NoHallu words, it shows that several models achieved strong performance, with SVM attaining the highest word-level performance and the attention-based model performing best at the sentence-level prediction. 
As expected, the cross-subject AUC scores are generally lower than the within-subject ones, likely due to inter-subject variability in EEG signals. Differences in brain anatomy, electrode placement, cognitive strategies, and noise make generalization across individuals more challenging~\citep{Apicella2024Toward}.
Another consistent trend is that sentence-level classification outperforms word-level classification. 
This is plausible because sentence-level prediction allows the model to integrate information across all constituent words, capturing contextual dependencies and cumulative signals.
The attention-based model in particular can exploit sequential dependencies via its internal weighting mechanism, which helps it better aggregate subtle signals across words.
The results indicate that the EEG signals we collected carry meaningful information for predicting hallucinated vs non-hallucinated content.
We further experimented by comparing HalluWrong vs NoHallu and HalluCorrect vs HalluWrong words. 
The results show that model performance drops significantly and does not exceed random chance by a meaningful margin.
This indicates that EEG signals contain discriminative information only when participants correctly recognize hallucinations.
When participants fail to detect hallucinated content, the model is unable to distinguish hallucinated words from non-hallucinated words based on EEG signals.

Overall, our experiments validate that, at both the word and sentence levels, EEG is a viable implicit feedback signal for detecting hallucination in generated content.
However, this prediction is reliable only when participants correctly recognize hallucination.~(addressing \textbf{RQ3})

\section{Conclusion}

In summary, this paper makes the following three contributions.
1)  We collected and will release an EEG dataset from 27 participants, in which subjects viewed text generated by MLLM, including both hallucinated and non-hallucinated content.
2)  We performed ERP analyses to probe the neural mechanisms of human recognition of MLLM-generated hallucinations and found that early attention and perceptual processing, semantic-thematic integration, inferential reasoning, and memory retrieval are all involved at very fine temporal resolution. 
Crucially, ERP differences between hallucination vs non-hallucination words only emerge when participants correctly recognize hallucination content, indicating that 
endogenous cognitive mechanisms that attenuate conflict detection when the hallucinated content appears fluent and contextually plausible.
3)  We demonstrated that it is possible to predict whether content contains hallucinations with EEG at both the word-level and sentence-level, but reliable prediction depends on correct recognition of hallucinated content by participants.
    

Despite the promising findings, this study has several limitations that must be acknowledged.
1) Although we have a relatively large number of participants (n = 27), which helps statistical reliability, each participant in our dataset viewed relatively few hallucination words, since the EEG data collection equipment is not portable and the sessions are time-consuming. 
2) Our setup was constrained to a laboratory setting. We made efforts to approximate real-world conditions, but there remains a gap between them. 
Factors such as ambient noise, participant movement, multitasking, natural reading behavior, and variations in attention in real life are not fully captured.

Our empirical findings suggest that the detection of hallucinated content involves memory retrieval and semantic matching.
Those cognitive processes may depend on whether participants have relevant knowledge. 
Future studies would be valuable to conduct experiments within groups possessing specialized backgrounds (e.g., experts in medicine, law, science) to assess how prior knowledge modulates EEG signatures of hallucination recognition. 
Although our study examined several categories of hallucination (relation, entity, attribute), a more fine‐grained investigation is needed to understand how different kinds of semantic and perceptual violations produce distinct neural effects. 
This would help map which categories are most difficult to detect, in which brain regions, and at what latencies, thereby informing both cognitive theory and model design.
Another promising direction is to move beyond offline analysis toward real-time or adaptive human–AI interaction systems. 
Future work could explore whether EEG-based signals of hallucination recognition can be decoded online and fed back to user interfaces in real time, enabling dynamic warning mechanisms, adaptive response generation, or confidence calibration.



\section*{Impact Statement}

This work aims to advance understanding of human–AI interaction and the neural mechanisms underlying the recognition of AI-generated hallucinations. 
The user study conducted in this work was approved by the Ethics Review Committee of the Artificial Intelligence Committee at Tsinghua University.
All experiments involving human participants were conducted under formal ethical approval, with procedures designed to ensure participant safety, informed consent, and protection of privacy. 
We believe this study poses minimal risk to participants and contributes positively to the development of safer and more trustworthy AI systems by providing insights into how humans perceive and evaluate potentially misleading AI-generated content.

\bibliography{citation}
\bibliographystyle{icml2026}

\newpage
\appendix
\onecolumn

\newpage
\section{Appendix}

\subsection{GPT4-HDM Verification}
\label{sec_verification}

We further validated the selected textual stimuli using the GPT4-HDM verification method, with a carefully designed prompt adapted from prior hallucination-detection settings. Specifically, we employed the following prompt to evaluate whether each text span contained non-factual or hallucinated information:

\begin{quote}
Given the following text span, your objective is to determine if the provided text contains non-factual or hallucinated information. You SHOULD give your judgment based on the world knowledge.

Text span: \textbf{[Provided Text]}

Now, determine if the above text span contains non-factual or hallucinated information. The answer you give MUST be ``Yes'' or ``No''.
\end{quote}

\subsection{ERP Analysis}
\label{appendix_erp}

\begin{table}[th]
  \centering
  \caption{Raw and FDR-corrected p-values (HalluCorrect vs. NoHallu words)}
  \label{tab_pvalues_h_n}
\begin{tabular}{lccccccc}
\toprule
 & pre-frontal & frontal & central & l-temporal & r-temporal & parietal & occipital \\
\midrule 
\multicolumn{8}{l}{\textbf{Raw p-values}} \\
50--120 ms & 0.0609 & 0.2282 & 0.6916 & 0.1084 & 0.0082 & 0.0093 & 0.8691 \\
120--280 ms & 0.0070 & 0.0002 & 0.0006 & 0.0009 & 0.0493 & 0.0731 & 0.0025 \\
280--550 ms & 0.1148 & 0.0849 & 0.0004 & 0.0379 & 0.0092 & 0.5621 & 0.0304 \\
550--750 ms & 0.0226 & 0.0121 & 0.0000 & 0.0109 & 0.0086 & 0.1085 & 0.0166 \\
\midrule 
\multicolumn{8}{l}{\textbf{FDR-corrected p-values}} \\
50--120 ms  & 0.1422 & 0.3194 & 0.8069 & 0.1897 & 0.0326 & 0.0326 & 0.8691 \\
120--280 ms & 0.0098 & 0.0016 & 0.0021 & 0.0021 & 0.0575 & 0.0731 & 0.0043 \\
280--550 ms & 0.1339 & 0.1189 & 0.0028 & 0.0663 & 0.0321 & 0.5621 & 0.0663 \\
550--750 ms & 0.0264 & 0.0213 & 0.0000 & 0.0213 & 0.0213 & 0.1085 & 0.0232 \\
\bottomrule
\end{tabular}
\end{table}

Tables~\ref{tab_pvalues_h_n} present the p-values before and after FDR correction($\alpha = 0.05$), for different ERP components across brain regions, comparing HalluWrong vs. NoHallu words.

\begin{table}[th]
  \centering
  \caption{Raw and FDR-corrected p-values (HalluCorrect vs. HalluWrong words)}
  \label{tab_pvalues_h_w}
\begin{tabular}{lccccccc}
\toprule
 & pre-frontal & frontal & central & l-temporal & r-temporal & parietal & occipital \\
\midrule 
\multicolumn{8}{l}{\textbf{Raw p-values}} \\
50--120 ms & 0.1616& 0.3523& 0.0973& 0.1889& 0.0937& 0.6133& 0.0056\\
120--280 ms & 0.1521& 0.0455& 0.0026& 0.0718& 0.0631& 0.0017& 0.2117\\
280--550 ms & 0.0360& 0.0293& 0.0346& 0.2912& 0.0274& 0.0408& 0.1306\\
550--750 ms & 0.0413& 0.0459& 0.0189& 0.0619& 0.0382& 0.0055& 0.0668\\
\midrule 
\multicolumn{8}{l}{\textbf{FDR-corrected p-values}} \\
50--120 ms  & 0.2644& 0.411& 0.227& 0.2644& 0.227& 0.6133& 0.0391\\
120--280 ms & 0.1775& 0.1005& 0.009& 0.1005& 0.1005& 0.009& 0.2117\\
280--550 ms & 0.0571& 0.0571& 0.0571& 0.2912& 0.0571& 0.0571& 0.1524\\
550--750 ms & 0.0643& 0.0643& 0.0643& 0.0668& 0.0643& 0.0382& 0.0668\\
\bottomrule
\end{tabular}
\end{table}

Tables~\ref{tab_pvalues_h_w} present the p-values before and after FDR correction($\alpha = 0.05$), for different ERP components across brain regions, comparing HalluWrong vs. HalluWrong words.

\begin{table}[th]
  \centering
  \caption{Raw and FDR-corrected p-values (HalluWrong vs. NoHallu words)}
  \label{tab_pvalues_w_n}
\begin{tabular}{lccccccc}
\toprule
 & pre-frontal & frontal & central & l-temporal & r-temporal & parietal & occipital \\
\midrule 
\multicolumn{8}{l}{\textbf{Raw p-values}} \\
50-120 ms  & 0.3095      & 0.1499  & 0.7489  & 0.7816     & 0.4652     & 0.2290   & 0.0801    \\
120-280 ms & 0.2067      & 0.0775  & 0.4556  & 0.2885     & 0.8881     & 0.0796   & 0.0750    \\
280-550 ms & 0.8963      & 0.8553  & 0.5562  & 0.5928     & 0.6642     & 0.4844   & 0.0958    \\
550-750 ms & 0.8155      & 0.9407  & 0.2379  & 0.7053     & 0.6984     & 0.0924   & 0.3060   \\
\midrule 
\multicolumn{8}{l}{\textbf{FDR-corrected p-values}} \\
50--120 ms  & 0.5416 & 0.5247 & 0.7816 & 0.7816 & 0.6513 & 0.5343 & 0.5247 \\
120--280 ms & 0.3617 & 0.1857 & 0.5315 & 0.4039 & 0.8881 & 0.1857 & 0.1857 \\
280--550 ms & 0.8963 & 0.8963 & 0.8963 & 0.8963 & 0.8963 & 0.8963 & 0.6706 \\
550--750 ms & 0.9407 & 0.9407 & 0.7140 & 0.9407 & 0.9407 & 0.6468 & 0.7140 \\
\bottomrule
\end{tabular}
\end{table}

\begin{table*}[th]
\caption{The statistical significance test results (F score) for different ERP components across brain regions for HalluWrong vs. NoHallu words.}
\label{tab_erp_anova_f_n_w}
\begin{center}
\begin{tabular}{cccccccc}
\toprule
F[1,26] & pre-frontal & frontal & central & l-temporal & r-temporal & parietal & occipital \\
\midrule
50-120 ms   & 1.0762      & 2.2069  & 0.1047  & 0.0786     & 0.5500     & 1.5207   & 3.3269    \\
120-280 ms  & 1.6805      & 3.3903  & 0.5743  & 1.1759     & 0.0202     & 3.3387   & 3.6863    \\
280-550 ms  & 0.0173      & 0.0340  & 0.3558  & 0.2935     & 0.1930     & 0.5038   & 2.9955    \\
550-750 ms  & 0.0556      & 0.0056  & 1.4622  & 0.1463     & 0.1536     & 3.0617   & 1.0923   \\
\bottomrule
\end{tabular}
\end{center}
\end{table*}

Tables~\ref{tab_erp_anova_f_n_w} and Table~\ref{tab_pvalues_w_n} present the statistical significance test results (F scores and p values before and after FDR correction($\alpha = 0.05$), respectively) for different ERP components across brain regions, comparing HalluWrong vs. NoHallu words. 
The results indicate that for all ERP components and in all examined regions of interest, the differences between HalluWrong and NoHallu words are not statistically significant.

\subsection{More Statistic Analysis}


\begin{table}[h]
\caption{Number of Correct Items per Participant}
\label{tab_correct_counts}
\centering
\small
\begin{tabular}{cc|cc|cc}
\hline
\textbf{ID} & \textbf{Correct} & \textbf{ID} & \textbf{Correct} & \textbf{ID} & \textbf{Correct} \\ \hline
P01 & 100 & P10 & 103 & P19 & 104 \\
P02 & 85  & P11 & 112 & P20 & 105 \\
P03 & 106 & P12 & 103 & P21 & 102 \\
P04 & 106 & P13 & 90  & P22 & 107 \\
P05 & 98  & P14 & 103 & P23 & 106 \\
P06 & 100 & P15 & 106 & P24 & 101 \\
P07 & 93  & P16 & 96  & P25 & 102 \\
P08 & 97  & P17 & 107 & P26 & 88  \\
P09 & 112 & P18 & 98  & P27 & 111 \\ \hline
\end{tabular}
\end{table}

Table~\ref{tab_correct_counts} presents the distribution of the number of correct items answered by all participants, where the mean is approximately 101.14, the standard deviation is 6.53, and the coefficient of variation is 6.46\% — reflecting a relatively high level of consistency in the number of correct items among the participants.

\subsection{Models}
\label{sec_model}

Selecting features at the ROI level, as we do in the main text, rather than individual electrodes, is a common practice in EEG research, as it helps reduce spurious correlations and improves robustness across subjects and trials. Following this standard approach, our feature definition is based on predefined ROIs and ERP time windows, rather than fine-grained, data-driven selection tied to specific electrodes. To further address the concern about optimistic bias, we conducted additional experiments where feature selection was performed independently within each training fold. The results show that the selected features are highly consistent across folds, with the same major ROIs (e.g., central, temporal, and occipital) repeatedly identified. Moreover, the corresponding classification performance shows no significant difference compared to our original setup. This provides strong empirical evidence that our results are not driven by data leakage or overfitting. From a cognitive neuroscience perspective, this consistency is also expected. The selected ROIs and time windows align well with established neural substrates of visual and language processing, and ROI-level aggregation is generally less sensitive to random noise or electrode-specific variability than single-electrode selection. Therefore, our feature design reflects both empirical robustness and neuroscientific priors, supporting its validity and generalizability.

The model structures and hyperparameters are as follows. For all models, the input features first undergo a preprocessing pipeline, which includes mean imputation for any missing values, followed by standard scaling to normalize the data.

{SVM (Support Vector Machine)} We use a Radial Basis Function (RBF) kernel. The regularization parameter $C$ is set to $1$. The model is configured to output probability estimates for classification.

{RF (Random Forest)} We set the number of trees in the forest to $100$. All other parameters are kept at their default values as specified in the scikit-learn library.

{GBDT (Gradient Boosting Decision Trees)} We set the number of boosting stages to $100$ and the learning rate to $0.1$. All other parameters are set to their default values.

{MLP (Multi-Layer Perceptron)} We implement a network using PyTorch. The architecture consists of a single hidden layer with $100$ units, which uses a ReLU activation function. A dropout layer with a probability of $0.5$ is applied after the activation function for regularization. The output layer is a linear layer that maps to the two output classes.

{Attention-based model} We use a Transformer Encoder architecture implemented in PyTorch. The input features are first projected into an embedding space with a dimension of $128$. This is followed by a 2-layer Transformer Encoder. Each encoder layer utilizes a multi-head attention mechanism with $8$ attention heads and a dropout rate of $0.5$. A final linear layer maps the encoder's output to the class scores.

{Training Configuration for Deep Models} For both deep learning models (MLP and Attention-based), we use the cross-entropy loss function and the Adam optimizer with a learning rate of $10^{-3}$. The models are trained for $300$ epochs with a batch size of $32$.

\subsection{More Results}

\begin{table}[t]
\caption{The recall results of word-level and sentence-level prediction. † indicates the result is significantly different with p-value\textless0.05 compared to the best model. * indicates including HalluWrong words in the training.}
\label{tab_experiments_recall}
\begin{center}
\begin{tabular}{lccccc}
\toprule
 \multirow{2}{*}{Settings}&\multirow{2}{*}{Models} & \multicolumn{2}{c}{word-level} & \multicolumn{2}{c}{sentence-level} \\
                         && $Recall_{within}$& $Recall_{cross}$& $Recall_{within}$& $Recall_{cross}$\\
\midrule
 \multirow{5}{*}{\makecell[c]{HalluCorrect\\vs\\NoHallu}}&SVM                     &     0.5740†&   0.3669†&       0.6610†&     0.4655†\\
 &RF                      &     0.2629†&   0.1169†&       0.0404†&     0.0162†\\
 &GBDT                    &     0.4104†&   0.2132†&       0.4602†&     0.0346†\\
 &MLP                     &     0.6141†&   0.3487†&       0.8952†&     \textbf{0.8244}\\
 &attention               &     \textbf{0.6617}&   \textbf{0.4421}&       \textbf{0.9310}&     0.7407†\\
\midrule
 \multirow{5}{*}{\makecell[c]{HalluWrong\\vs\\NoHallu}}&SVM                     &     0.0000&   0.0000&       0.0023&     0.0000\\
 & RF                      & 0.0000& 0.0000& 0.0000&0.0000\\
 & GBDT                    & 0.0023& 0.0792& 0.1201&0.0310\\
 & MLP                     & 0.0000& 0.0028& 0.0471&0.0523\\
 & attention               & 0.0000& 0.0000& 0.0847&0.0000\\
 \midrule
 \multirow{5}{*}{\makecell[c]{HalluCorrect\\vs\\HalluWrong}}&SVM                     &     1&   1&       1&     1\\
 & RF                      & 0.9886& 0.9641& 0.998&0.9941\\
 & GBDT                    & 0.8864& 0.9589& 0.9663&0.9609\\
 & MLP                     & 0.9368& 0.9605& 0.9888&0.9985\\
 & attention               & 0.8556& 0.9079& 0.9832&0.9806\\
\bottomrule
\end{tabular}
\end{center}
\end{table}

Table~\ref{tab_experiments_recall} shows the more results of word-level and sentence-level prediction.

\begin{table}[t]
\centering
\caption{Comparison between EEG-based hallucination detection and representative AI-based LVLM hallucination detection methods.}
\label{tab:hallu_detection_compare}
\begin{tabular}{lcc}
\toprule
\textbf{Method} & \textbf{Type} & \textbf{AUC} \\
\midrule
EEG-based (cross-subject, attention, ours) & Neural-based & 0.931 \\
Uncertainty-based (PPL) & Uncertainty-based & 0.876 \\
Uncertainty-based (Token Confidence) & Uncertainty-based & 0.892 \\
Confidence-based (Consistency via re-asking) & Confidence-based & 0.881 \\
Self-verification (Verification prompt) & Self-verification & 0.851 \\
\bottomrule
\end{tabular}
\end{table}

To compare our prediction results with existing LVLM hallucination detection approaches, we evaluated three representative AI-based hallucination detection methods alongside our EEG-based detection framework on the constructed hallucination dataset, as shown in Table~\ref{tab:hallu_detection_compare}. The compared methods include uncertainty-based detection, confidence-based consistency checking, and self-verification approaches.
We found that the EEG-based detection method consistently achieved higher AUC scores than these conventional model-based approaches. We believe this advantage may arise from two main factors. First, our EEG signals were derived from the averaged event-related potential (ERP) responses of 27 participants, providing a more stable and less noisy representation of human cognitive processing. Second, neural signals offer an intrinsic process-level measurement that is not restricted to post-hoc behavioral outputs or heuristic model-internal statistics. As a result, EEG signals may capture certain aspects of hallucination processing that are inaccessible to standard model-based detection methods.




\end{document}